\documentclass{article}
\usepackage{main}

\usepackage{times}
\usepackage{xcolor}
\usepackage{amsmath}
\usepackage{graphicx}
\usepackage{graphics}
\usepackage{xcolor,colortbl}
\usepackage{amssymb}
\usepackage[]{algorithm2e}
\usepackage{soul}
\usepackage[utf8]{inputenc}
\usepackage[small]{caption}
\usepackage[belowskip=0pt,aboveskip=5pt]{caption}

\title{Submodularity-Inspired Data Selection for Goal-Oriented Chatbot Training Based on Sentence Embeddings}

\iftrue
\author{
Mladen Dimovski$^{1}$,
Claudiu Musat$^{2}$, 
Vladimir Ilievski$^{1}$,
Andreea Hossmann$^{2}$,
Michael Baeriswyl$^{2}$
\\ 
$^{1}$ School of Computer and Communication Sciences, EPFL, Switzerland\\
$^{2}$ Artificial Intelligence Group - Swisscom AG\\
firstname.lastname@\{epfl.ch, swisscom.com\}
}

\fi

\begin{document}

\maketitle

\begin{abstract}
Spoken language understanding (SLU) systems, such as goal-oriented chatbots or personal assistants, rely on an initial natural language understanding (NLU) module to determine the intent and to extract the relevant information from the user queries they take as input. SLU systems usually help users to solve problems in relatively narrow domains and require a large amount of in-domain training data.
This leads to significant data availability issues that inhibit the development of successful systems.
To alleviate this problem, we propose a technique of data selection in the low-data regime that enables us to train with fewer labeled sentences, thus smaller labelling costs.

We propose a submodularity-inspired data ranking function, the \textit{ratio-penalty marginal gain}, for selecting data points to label based only on the information extracted from the textual embedding space. We show that the distances in the embedding space are a viable source of information that can be used for data selection. Our method outperforms two known active learning techniques and enables  cost-efficient training of the NLU unit. Moreover, our proposed selection technique does not need the model to be retrained in between the selection steps, making it time efficient as well.
\end{abstract}

\section{Introduction}

In their most useful form, goal-oriented dialogue systems need to understand the user's need in great detail.  
A typical way of structuring this understanding is the separation of intents and slots that can be seen as parameters of the intents.
Slot filling, also known as entity extraction, is  finding the relevant information in a user query that is needed for its further processing by the SLU system. For example, in a restaurant reservation scenario, given the sentence \linebreak \textit{Are there any French restaurants in downtown Toronto?} as an input, the task is to correctly output, or fill, the following slots: \textit{\{cuisine: French\}} and \textit{\{location: downtown Toronto\}}.
\vspace{0.2cm}

The slot filling task is usually seen as a sequence tagging problem where the goal is to tag each relevant word token with the corresponding slot name using the B--I--O (Begin, Inside, Outside) convention. The table below shows how we would correctly tag the previous example sentence.
\vspace{0.2cm}
\begin{center}
    \begin{tabular}{ | p{1.7cm} | p{1.7cm}  | p{1.7cm}  | p{1.7cm}  |  }
    \hline
    Are \cellcolor{gray!25}& there\cellcolor{gray!25} & any \cellcolor{gray!25} & French\cellcolor{gray!25} \\ \hline
    O & O & O & B-Cuisine \\ \hline
    \end{tabular}
    \begin{tabular}{ | p{1.7cm} | p{1.7cm}  | p{1.7cm}  | p{1.7cm}  |  }
    \hline
    restaurants\cellcolor{gray!25} & in \cellcolor{gray!25}& downtown \cellcolor{gray!25} & Toronto\cellcolor{gray!25} \\ \hline
    O & O & B-Location & I-Location \\ \hline
    \end{tabular}
\end{center}
\vspace{0.3cm}

 Most methods created for slot filling are supervised and  require large amounts of labeled in-domain sentences in order to perform well. However, it is usually the case that only very little or no training data is available. Annotating new sentences is an expensive process that requires considerable human effort and, as a result, achieving good performance with as little data as possible becomes an important concern.
\vspace{0.2cm} 

 Our solution to the data availability problem relies on a better way of selecting the training samples to be labeled. If a limited amount of resources are available, we want to enable the user to spend them in the most efficient way. More precisely, we propose a method for \textit{ranking} the unlabeled sentences according to their utility. We measure the latter with the help of a ranking function that satisfies the principles of submodularity, a concept known to capture the intuition present in data selection. We run experiments on three different publicly available slot filling datasets: the MIT Restaurant, the MIT Movie and the ATIS datasets \cite{liu2013asgard,liu2013query}. We are interested in measuring the model's performance when it is trained with only a \textit{few dozen} labeled sentences, a situation we refer to as \textit{low-data regime}. We compare our proposed selection method to several standard baselines, including two variants of active learning.
\vspace{0.2cm}

 We identify our three main contributions:
\begin{itemize}
\item We show that the space of raw, unlabeled sentences contains information that we can use to choose the sentences to label.
\item We create a submodularity-inspired ranking function for selecting the potentially most useful sentences to label.
\item We apply this data selection method to the problem of slot filling and prove that the model's performance can be considerably better when the training samples are chosen in an intelligent way.
\end{itemize}

 In Section \ref{sec:dataselection}, we provide some background and details about the novel data selection technique. In Section \ref{sec:experiments}, we describe the datasets used in our experiments and the baselines that we use as a comparison reference point. Finally, in Section \ref{sec:results}, we present and discuss the obtained results.

\section{Related Work}
\subsection{Slot Filling}
Numerous models have been proposed to tackle the slot filling task, of which a comprehensive review was written by \cite{mesnil2015using}. However, the most successful methods that have emerged are neural network architectures and, in particular, recurrent neural network schemes based on word embeddings \cite{kurata2016leveraging,ma2016end,liu2016attention,zhang2016joint,zhu2017encoder,zhai2017neural}. The number of proposed model variants in the recent literature is abundant, with architectures ranging from encoder-decoder models \cite{kurata2016leveraging}, models tying the slot filling problem with the closely related intent detection task \cite{zhang2016joint} and even models that  use the attention mechanism \cite{liu2016attention}, originally introduced in \cite{bahdanau2014neural}.  The final model that we adopted in our study is a bi-directional LSTM network that uses character-sensitive word embeddings, together with a fully-connected dense and linear CRF layer on top \cite{huang2015bidirectional,lample2016neural}. We provide the model's specifications and more details in the appendix.

\subsection{Low-Data Regime Challenges}

Typically, machine learning models work reasonably well when trained with a sufficient amount of data: for example, reported results for the popular ATIS domain benchmark go beyond 95\% F1 score \cite{liu2016attention,zhang2016joint}. However, performance significantly degrades when little amount of training data is available, which is a common scenario when a new domain of user queries is introduced.  There are two major approaches used to handle the challenges presented by the scarcity of training data:
\begin{itemize}
    \item The first strategy is to train a multi-task model whose purpose is to deliver better performance on the new domain by using patterns learned from other closely related domains for which sufficient training data exists \cite{jaech2016domain,hakkani2016multi}
    \item The second strategy is to select the few training instances that we can afford to label. One well known approach are the active learning strategies \cite{fu2013survey,angeli2014stanford} that identify data points that are close to the separation manifold of an imperfectly trained model.
\end{itemize}

In our work, we focus on the latter scenario, as experience has shown that high-quality in-domain data is difficult to replace by using other techniques. Therefore, we assume that we can choose the sentences we want to label and the main question is how to make this choice in a way that would yield the model's best performance.

\section{Data Selection}
\label{sec:dataselection}
\subsection{Submodularity and Rankings}
Let $V$ be a ground set of elements and ${F: 2^V \to \mathbb{R}}$ a function that assigns a real value to each \textit{subset} of $V$. $F$ is called \textit{submodular} if the incremental benefit of adding an element to a set diminishes as the context in which it is considered grows. Formally, let $X$ and $Y$ be subsets of $V$, with ${X\subseteq Y\subseteq V}$. Then, $F$ is submodular if for every $e\in V\setminus Y$, $F(e|Y)  \leq F(e|X)$ where ${F(e|A) = F(\{e\} \cup A)-F(A)}$ is the benefit, or the \textit{marginal gain} of adding the element $e$ to the set $A$. The concept of submodularity captures the idea of diminishing returns that is inherently present in data selection for training machine learning models. 
 In the case of the slot filling problem, the ground set $V$ is the \textit{set of all available unlabeled sentences} in a dataset and the value $F(X)$ for a set $X\subseteq V$ is a score measuring the utility of the sentences in $X$.
Submodular functions have already been successfully used in document summarization \cite{lin2011class,lin2012learning}, and in various tasks of data subset selection \cite{kirchhoff2014submodularity,wei2014submodular,wei2015submodularity}. However, to the best of our knowledge, they have not yet been studied in the slot filling context.
\vspace*{0.2cm}

 An important hypothesis that is made when submodular functions are used for data selection is that if $X$ and $Y$ are two sets of data points for which $F(X)\leq F(Y)$, then using $Y$ for training would give an overall better performance than using $X$. If we have a predefined size for our training set, i.e., a maximum number of samples that we can afford to label, then we would need to find the set $X$ that maximizes $F(X)$ with a constraint on $|X|$. Cardinality-constrained submodular function maximization is an NP-hard problem and we usually have to resort to greedy maximization techniques (see \cite{krause2014submodular}). If the function is monotone, then the greedy solution, in which we iteratively add the element with the largest marginal gain with respect to the already chosen  ones, gives a $(1-1/e)$-approximation guarantee \cite{nemhauser1978analysis}. As a byproduct, this greedy procedure also produces a \textit{ranking} of the $n=|V|$ sentences, i.e., outputs a ranking permutation $\pi : [n] \to [n]$ that potentially orders them according to their usefulness. Therefore, a monotone submodular function $F$ naturally defines a selection criteria $\pi_F$ that is the order in which the elements are chosen with the greedy optimization procedure. In our work, we explore different data selection criteria, without limiting ourselves to properly defined submodular functions. 

\subsection{Sentence Similarity}
Unless we perform the selection of the sentences to label based on some intrinsic attributes such as their length, an important metric we need to define is the similarity between \textit{a pair} of sentences. 
The need to introduce such a measure of similarity appeared simultaneously with the development of active learning techniques for text classification. For example, in \cite{mccallum1998employing}, the similarity between two documents $x$ and $y$, their analogue to our sentences, is measured by an exponentiated Kullback-Leibler divergence between the word occurrence empirical measures ${\hat{\mathbb{P}}(W=w|x)}$ and ${\lambda \hat{\mathbb{P}}(W=w|y)+(1-\lambda)\hat{\mathbb{P}}(W=w)}$ where $\lambda$ is a smoothing parameter. In the recent literature, however, the focus has shifted to using word or sentence embeddings as a basis for almost all tasks involving text processing. The continuous (Euclidean) space embeddings have become fundamental building blocks for almost all state-of-the-art NLP applications.
\vspace*{0.2cm}

 In our study, we use a recently developed technique of producing sentence embeddings, \texttt{sent2vec} \cite{pagliardini2017unsupervised}. The \texttt{sent2vec} method produces continuous sentence representations that we use to define the similarity. More precisely, for every sentence $s \in V$, the \texttt{sent2vec} algorithm outputs a $700$-dimensional vector embedding ${e(s)\in \mathbb{R}^{700}}$ that we in turn use to define the similarity between a pair of sentences $x$ and $y$ as follows:
\begin{equation}
    \mbox{sim}(x,y) = \exp\big(-\beta \|e(x)-e(y)\|_2\big)
\end{equation}
where 
\begin{equation}
\beta = \bigg( \frac{1}{|V|(|V|-1)} \sum_{u \in V, w \in V} \|e(u)-e(w) \|_2 \bigg)^{-1}
\end{equation}
is a scaling constant, measuring the concentration of the cloud of points in the space. It is the inverse of the average distance between all pairs of embeddings. Note that $\beta$ will in general depend on the dataset, but it is not a hyper-parameter that needs to be tuned. Finally, the exponential function condenses the similarity to be in the interval $[0,1]$ as $\beta>0$. 
\vspace{0.2cm}

 For a fixed sentence $s$, the most similar candidates are those whose embedding vectors are the closest to $e(s)$ in the embedding space. Tables \ref{tab:sentences1} and \ref{tab:sentences3} present two example sentences (shown in bold), one from the MIT Restaurant and another from the MIT Movie dataset, together with their closest neighbors. Based on the numerous examples that we analyzed, it appears that the closeness in the embedding space is in line with the human perception of sentence similarity. Therefore, selecting a particular sentence for training should largely diminish the usefulness of its closest neighbors.
\vspace{0.2cm}

\begin{figure}
    \centering
    \includegraphics[width=\linewidth]{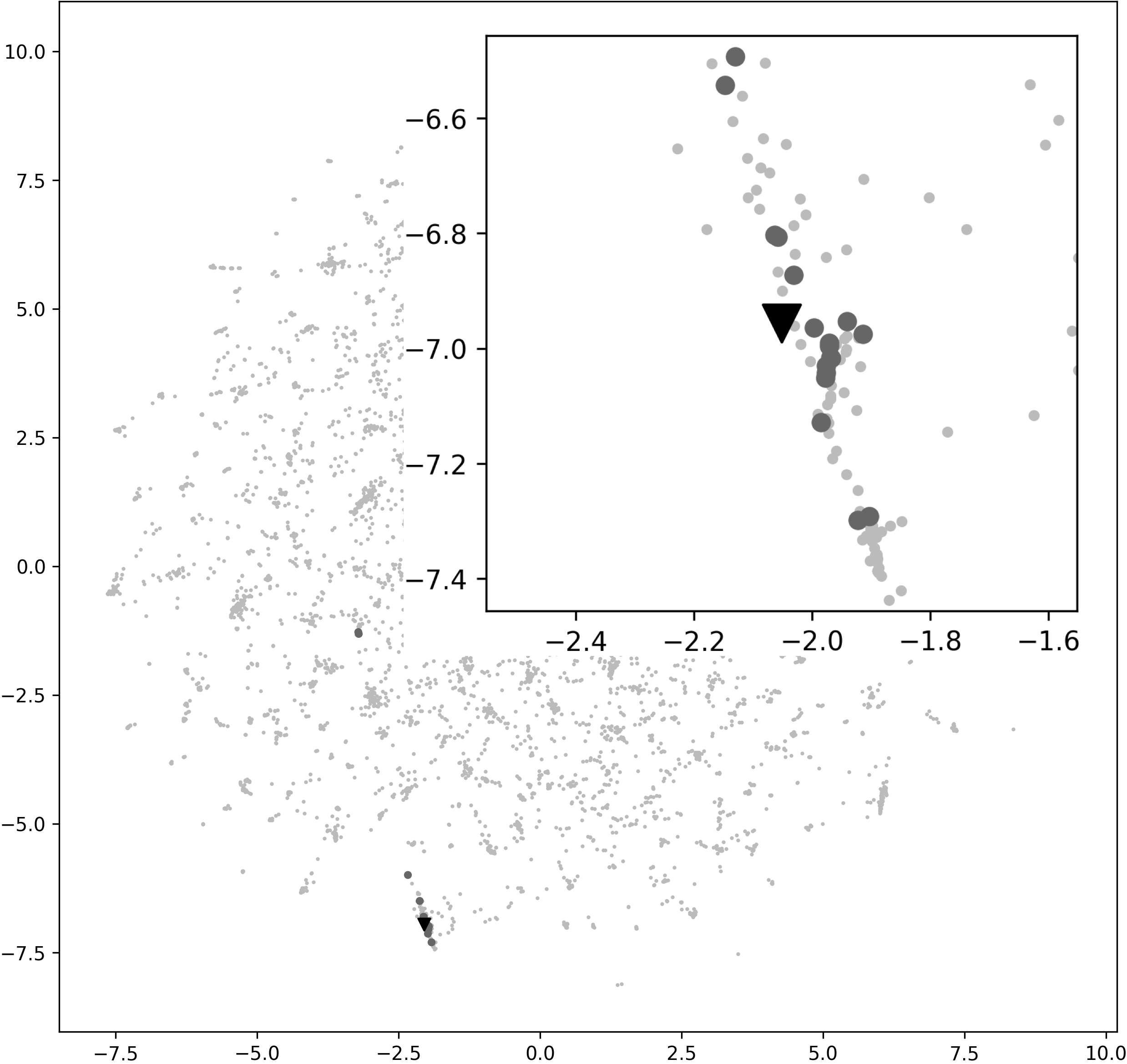}
    \caption{A $2$-dimensional t-SNE projection of the cloud of points representing the embeddings of the sentences of the MIT Movie domain. The overlapping plot gives a closer view of the small cluster at the bottom left; the corresponding sentences are shown in Table \ref{tab:sentences3}}
    \label{fig:cloud}
\end{figure}

\begin{table}[b]
\begin{center}
    \begin{tabular}{ | p{0.94\linewidth} |  }
    \hline
    \textbf{whats a cheap mexican restaurant here} \cellcolor{gray!25}  \\ \hline
 we are looking for a cheap mexican restaurant\\ \hline
 i am looking for a cheap mexican restaurant\\ \hline
 is ixtapa mexican restaurant considered cheap\\ \hline
    \end{tabular}
    \caption{\label{tab:sentences1}An example sentence $s$ from the MIT Restaurant domain and the sentences corresponding to the three closest points to $e(s)$}
\end{center}
\end{table}
\begin{table}[b]
\begin{center}
    \begin{tabular}{ | p{0.94\linewidth} |  }
    \hline
   \textbf{what was the movie that featured over the rainbow} \cellcolor{gray!25}  \\ \hline
find me the movie with the song over the rainbow\\ \hline
 what movie was the song somewhere out there featured in\\ \hline
what movie features the song hakuna matata\\ \hline
    \end{tabular}
    \caption{\label{tab:sentences3}An example sentence $s$ from the MIT Movie domain and the sentences corresponding to the three closest points to $e(s)$}
\end{center}
\end{table}
 A $2$-dimensional t-SNE projection of the cloud of points of the MIT Movie domain, together with an isolated example cluster, is shown on Figure \ref{fig:cloud}. The large black triangle in the cluster is an example sentence and the darker dots are its closest neighbors. Closeness is measured in a $700$-dimensional space and distances are not exactly preserved under the $2$-dimensional t-SNE projection portrayed here. The cluster shown on Figure \ref{fig:cloud} corresponds to the sentences in Table \ref{tab:sentences3}.

\subsection{Coverage Score}
\label{sec:cs}
Once we have defined a similarity metric between sentences, it can be used in the definition of various submodular functions. An important example is the \textit{coverage score} function that evaluates every subset of sentences $X$ in the following way  \cite{lin2011class}:
\vspace{0.1cm}
\begin{equation}
    C(X) = \sum_{x \in X} \sum_{y\in V} \mbox{sim}(x,y).
\end{equation}

Intuitively, the inner sum measures the total similarity of the sentence $x$ to the whole dataset; it is a score of how well $x$ covers $V$. The marginal gain of a sentence $s$ is given by
\begin{equation}
    C(s|X) = C(\{s\} \cup X) - C(X) = \sum_{y\in V}\mbox{sim}(s, y).
\end{equation}

\begin{figure}
    \centering
    \includegraphics[width=\linewidth]{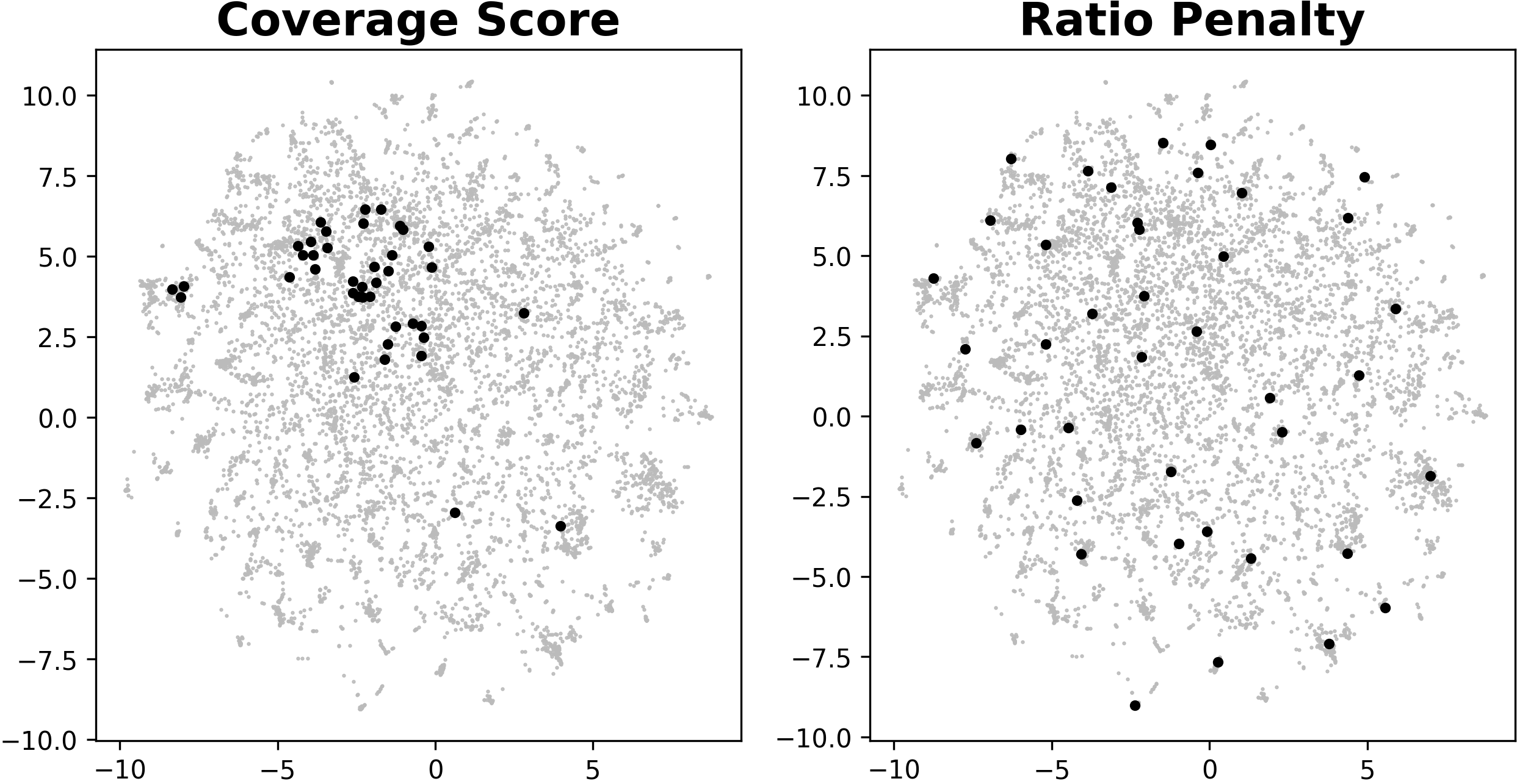}
    \caption{The position of the top forty points in the $2$-dimensional t-SNE projection of the cloud corresponding to the MIT Restaurant dataset according to two different rankings}
    \label{fig:clouds4}
\end{figure}

It can be easily seen that the marginal gain $C(s|X)$ does not depend on $X$, but only on $s$ itself. This makes the function $C$, strictly speaking, modular. The cardinality-constrained greedy optimization procedure for $C$ would, in this case, output the optimal solution. Table~\ref{tab:highestcs} presents the top three coverage score sentences from the MIT Restaurant dataset, and Figure~\ref{fig:clouds4} shows that these points tend to be centrally positioned in the cloud.
\vspace{0.2cm}

\begin{table}
\begin{center}
    \begin{tabular}{ | p{0.94\linewidth} |  }
    \hline
   \textbf{The top three sentences with highest coverage score} \cellcolor{gray!25}  \\ \hline
1. i need to find somewhere to eat something close by im really hungry can you see if theres any buffet style restaurants within five miles of my location\\ \hline
2. i am trying to find a restaurant nearby that is casual and you can sit outside to eat somewhere that has good appetizers and food that is a light lunch\\ \hline
3. i need to find a restaurant that serves sushi near 5 th street one that doesnt have a dress code\\ \hline
    \end{tabular}
    \caption{\label{tab:highestcs}The top three coverage score sentences from the MIT Restaurant domain}
\end{center}
\end{table}
 The coverage score function suffers from that it only considers how much a sentence is useful \textit{in general}, and not in the context of the other sentences in the set. Hence, it might pick candidates that cover much of the space, but could be very similar to each other. In order to deal with this problem, an additional penalty or diversity reward term is usually introduced and the marginal gain takes the form of
\begin{equation}
    D(s|X) =  \sum_{y \in V} \mbox{sim}(s, y) - \alpha \sum_{x \in X} \mbox{sim}(s,x)
\end{equation}
where $\alpha$ is a parameter that measures the trade-off between the coverage of $s$ and its similarity to the set of already chosen sentences $X$. The additional term makes the function $D$ submodular as $D(s|X)$ is decreasing in $X$. However, $D$ features an additional parameter that needs to be tuned and moreover, experiments with both $C$ and $D$ (see Figure \ref{fig:resresall} in the results section) yielded mediocre results. In the next section, we introduce our new selection method that features a non-linear penalization and does not have any additional parameters.

\subsection{Ratio-Penalty Marginal Gain}
 We propose a direct definition of a marginal gain of an element with respect to a set. This is an alternative to providing a submodular function for which we derive the marginal gain expression and then use it to find the maximizing set of a given size. We then use this marginal gain to rank and select the sentences that are likely the most valuable for labeling. We call it the \textit{ratio-penalty} marginal gain and define it as:
\begin{equation}
\label{rpmg}
    F(s|X) = \frac{\sum_{y \in V}\mbox{sim}(s,y)}{ 1 + \sum_{x \in X}\mbox{sim}(s,x)}.
\end{equation}

We cannot uniquely define a submodular function that generates the ratio-penalty marginal gain. To see this, notice that, for example, ${F(a|\emptyset) + F(b|\{a\}) \neq F(b|\emptyset) + F(a|\{b\})}$, which already makes the definition of $F(\{a,b\})$ ambiguous. Nevertheless, the above expression (\ref{rpmg}) satisfies the submodularity condition by being decreasing in $X$.
\vspace{0.2cm}

 The ratio-penalty marginal gain is again a trade-off between the coverage score of a sentence and its similarity to those already chosen. An important change is that the penalty is introduced by division, instead of by subtraction as in
$D(s|X)$ from the previous section.
To gain more intuition, notice that, as the logarithm function is increasing, the induced ranking $\pi_F$ is the same as the ranking $\pi_{\tilde{F}}$ produced by $\tilde{F}(s|X)$, where we define $\tilde{F}(s|X) = \log F(s|X)$, i.e.,
\begin{equation}
    \tilde{F}(s|X) = \log{\sum_{y \in V}\mbox{sim}(s,y)} - \log{\bigg( 1 + \sum_{x \in X}\mbox{sim}(s,x)\bigg)}.
\end{equation}

In summary, we start with the Euclidean distance between the embeddings of the sentences, we re-scale it by dividing it by the average distance $1/\beta$ and squash it exponentially. This defines our similarity metric. Then, we do an aggregate computation in this new space by summing the similarities, and we revert to the original scale by taking the logarithm.

\section{Experiments}
\label{sec:experiments}

\subsection{Datasets and Method}
We experiment with three different publicly available datasets whose details are shown in Table \ref{tab:domains}. Each one contains few thousand training samples, out of which we select, following some selection criteria $\pi$, only a few dozen.
More precisely, we are interested in the model's performance when it is trained only with $k=10, 20, 30,..., 100$ labeled sentences. This selection simulates the behaviour of a system that needs to be trained for a newly available domain. We measure the performance by the best achieved F1 score during training on a separate test set that we have for each domain. The final column of Table \ref{tab:domains} shows the performance of our adopted model trained on the \textit{full} datasets. These are the best results that we can hope to achieve when training with a proper subset of training samples.
We will denote by $\mathcal{T}_n=\{x_1, x_2,...,x_n\}$ the full training set of a particular domain, assuming that each $x_i$ is a (sentence, tags) pair. The training set comprising the subset of $k$ training samples that we select using the ranking $\pi$ will be denoted by $\mathcal{T}^{\pi}_k = \{ x_{\pi(1)},x_{\pi(2)},..., x_{\pi(k)}\}$.
\begin{table}
\begin{center}
    \begin{tabular}{ | p{2.25cm} | p{0.9cm}  | p{0.9cm}  | p{0.9cm}  |  p{1.3cm}|}
    \hline
    \textbf{Domain} \cellcolor{gray!25}& \textbf{\#train} \cellcolor{gray!25} & \textbf{\#test} \cellcolor{gray!25} & \textbf{\#slots}\cellcolor{gray!25} & \textbf{F1 score}\cellcolor{gray!25}  \\ \hline
    MIT Restaurant & 7661 & 1522 & 17 & 80.11 \\ \hline
    MIT Movie & 9776 & 2444 & 25 & 87.86\\ \hline
    ATIS & 4978 & 893 & 127 & 95.51\\ \hline
    \end{tabular}
    \caption{\label{tab:domains}The three domains and their basic information (number of training samples, number of samples in the test set and number of different slots). The MIT Restaurant and MIT Movie datasets contain user queries about restaurant and movie information. The Airline Travel Information Services (ATIS) dataset mainly contains questions about flight booking and transport information}
\end{center}
\end{table}
\subsection{Baselines}
To the best of our knowledge, data selection techniques have not yet been applied to the slot filling problem. As a result, there is no direct benchmark to which we can compare our method. Instead, we use three baselines that were used in similar situations: random data selection, classic active learning and an adaptation of a state-of-the-art selection method -- randomized active learning \cite{angeli2014stanford}. 

\subsubsection{Random Selection of Data}
To select $k$ sentences out of the total $n$, we uniformly pick a random permutation $\sigma \in \mathcal{S}_n$ and take as our training set $\mathcal{T}^{\sigma}_k = \{ x_{\sigma(1)}, x_{\sigma(2)}, ..., x_{\sigma(k)}\}$. 
Although random data selection occasionally picks irrelevant sentences, it guarantees diversity, thus leading to a reasonable performance. We show that some complex methods either fail to outperform the random baseline or they improve upon it only by a small margin.

\subsubsection{Classic Active Learning Framework}
For our second baseline, we choose the standard active learning selection procedure. We iteratively train the model and select new data based on the model's uncertainty about new unlabeled points. To calculate the uncertainty for a sentence $x=(w_1, w_2,..., w_k)$ containing $k$ word tokens, we adapt the least confidence (LC) measure \cite{fu2013survey,culotta2005reducing}. We define the uncertainty of a trained model $\mathbf{\Theta}$ for the sample $x$ as the average least confidence uncertainty across the labels
\begin{equation}
    u(x) = \frac{1}{k}\sum_{i=1}^k(1-p_{\mathbf{\Theta}}(\hat{y}_{w_i}|x))
\end{equation}
where $\hat{y}_w$ is the most likely tag for the word $w$ predicted by  $\mathbf{\Theta}$ and $p_{\mathbf{\Theta}}(\hat{y}_w|x)$ is its softmax-normalized score output by the dense layer of the model network.
\vspace{0.2cm}

 We proceed by selecting batches of ten sentences, thus performing a \textit{mini-batch adaptive active learning} \cite{wei2015submodularity} as follows. The first ten sentences $\mathcal{T}_{10}$ used for the initial training of the model $\mathbf{\Theta}$ are picked randomly. Then, at each iteration ${k\in [20, 30, 40,..., 100]}$, we pick a new batch of ten sentences $\mathcal{N}$ for which $\mathbf{\Theta}$ is the most uncertain and \textit{retrain} the model with the augmented training set $\mathcal{T}_{k} = \mathcal{T}_{k-10} \cup \mathcal{N}$. 
\vspace*{0.2cm}

\subsubsection{Randomized Active Learning}
 In the text processing scenario, the classic active learning algorithm has the drawback of choosing sentences that are not good representatives of the whole dataset. The model is often the least certain about samples that lie in sparsely populated regions and this blindness to the input space density often leads to a poor performance. To address this problem, data is usually selected by a weighted combination of the uncertainty about a sample and its correlation to the other samples \cite{fu2013survey,culotta2005reducing}. 
However, this approach requires finding a good correlation metric and also tuning a trade-off parameter of confidence versus representativeness. The latter is not applicable in our scenario because we would need to access a potential validation set, which deviates from our selection principle of labeling the least data possible. Instead, we adapt a well-performing technique proposed by \cite{angeli2014stanford} in which samples are selected randomly \textit{proportionally to the model's uncertainty}. More precisely, a sample sentence $x$ is selected with probability $u(x)/\sum_{y \in V} u(y)$.
Although uncertainty will again be the highest for the poor samples, as their number is small, they will contain only a tiny percent of the total uncertainty mass across the whole dataset. Consequently, they will have very little chance of being selected. 
\begin{figure}
    \centering
    \includegraphics[width=\linewidth]{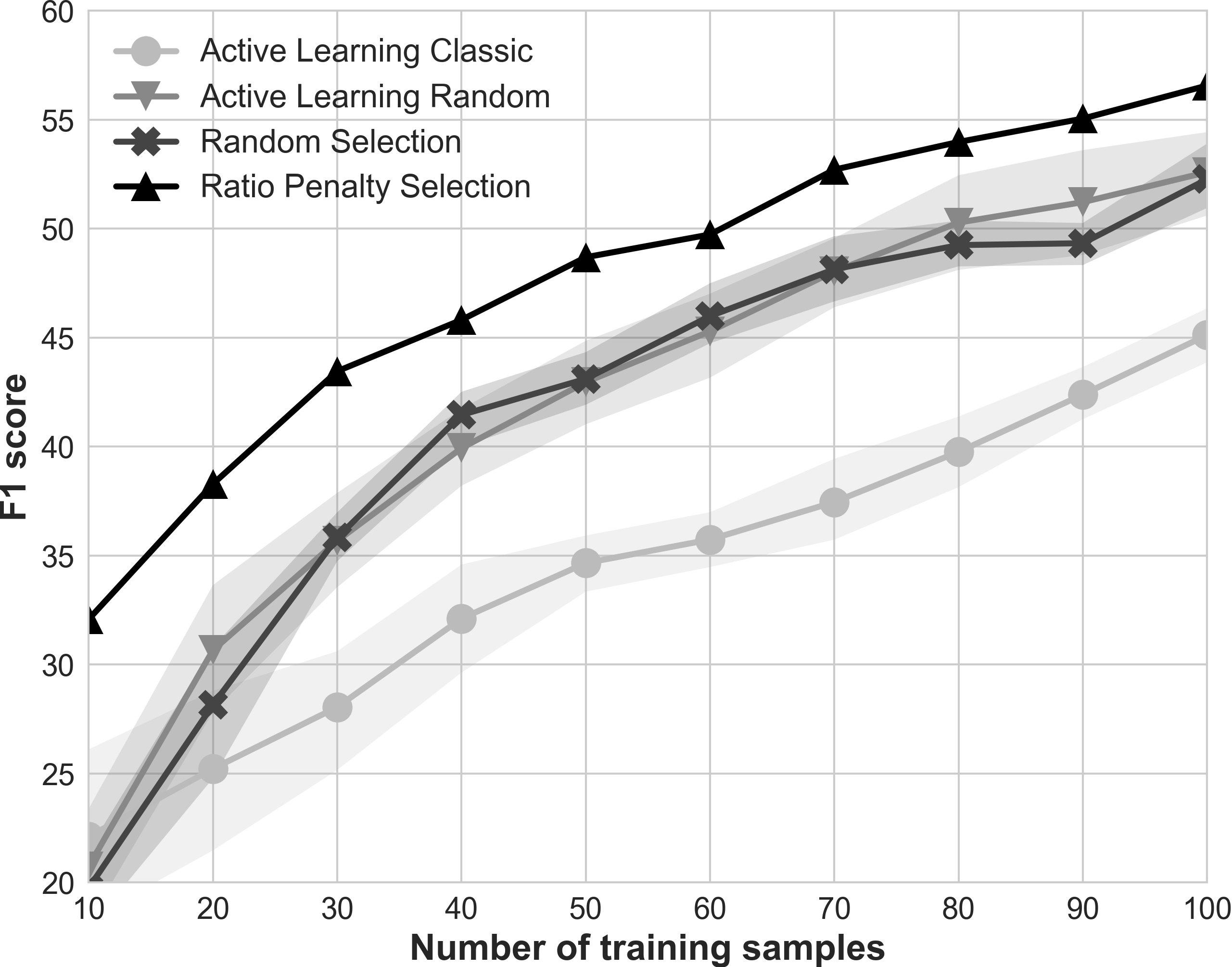}
    \caption{Comparison between the ratio-penalty and the three baseline selection methods for the MIT Restaurant dataset}
    \label{fig:resres}
\end{figure}
\begin{figure}
    \centering
    \includegraphics[width=\linewidth]{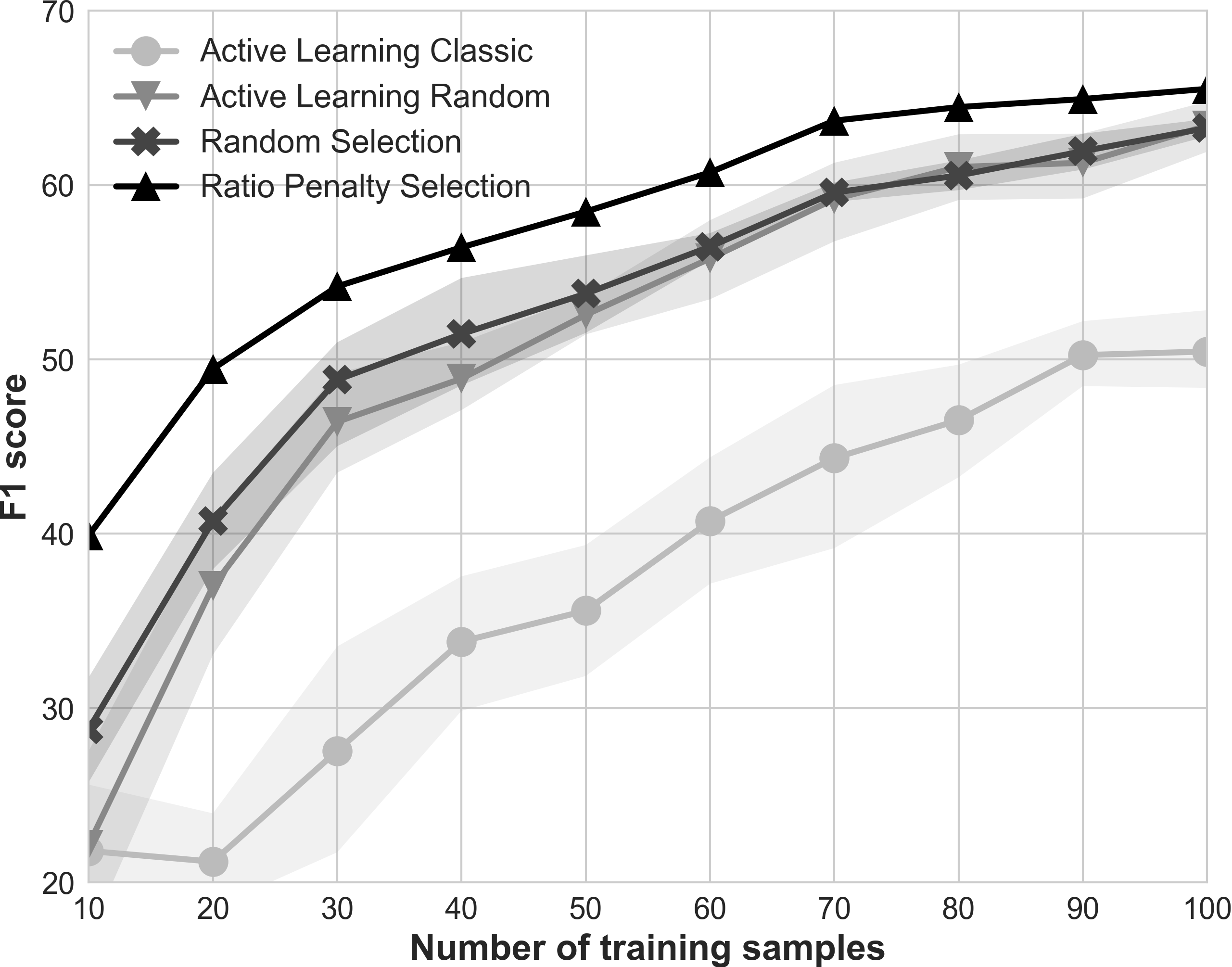}
    \caption{Comparison between the ratio-penalty and the three baseline selection methods for the MIT Movie dataset}
    \label{fig:resmov}
\end{figure}
\section{Results}
\label{sec:results}
\begin{figure}
    \centering
    \includegraphics[width=\linewidth]{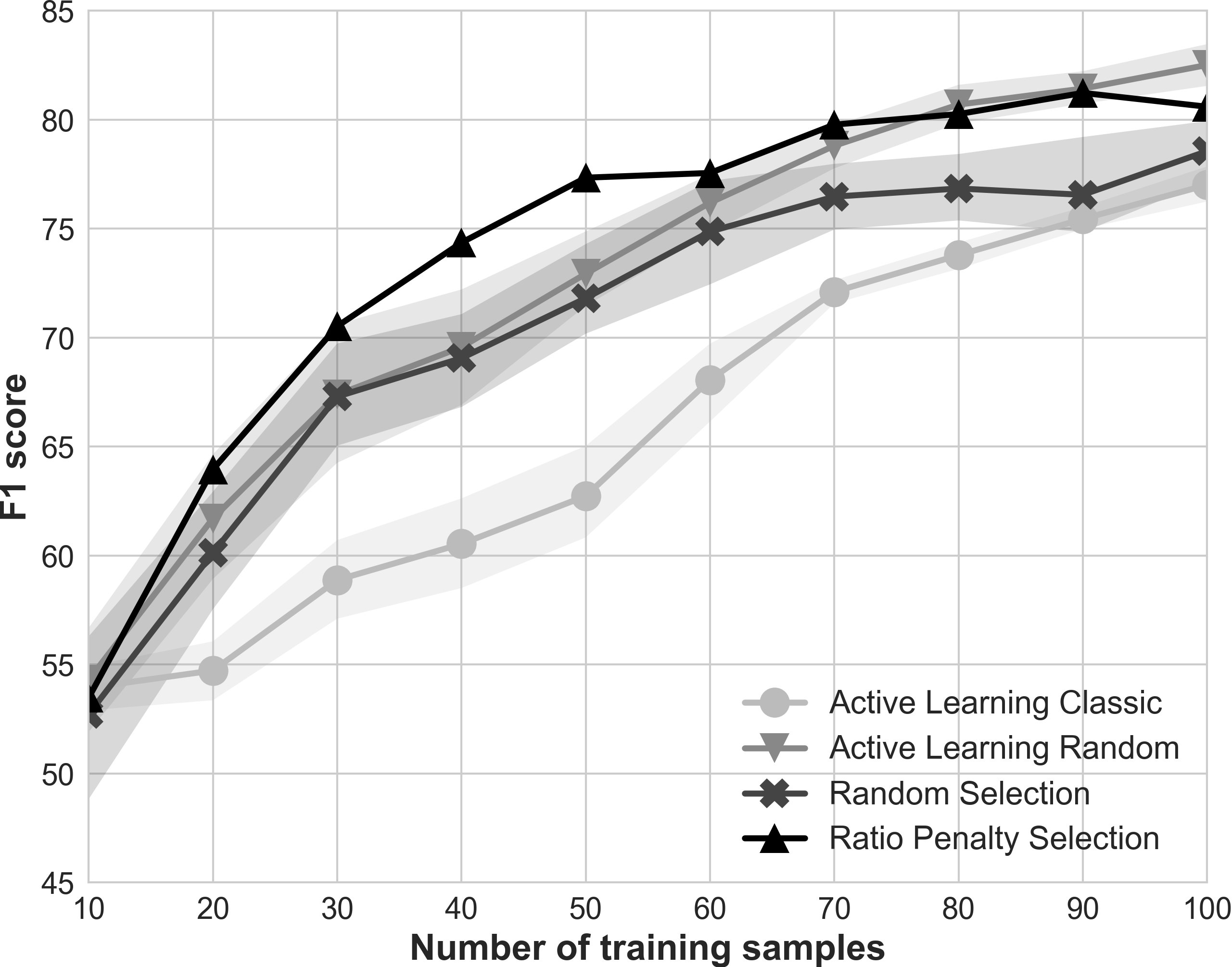}
    \caption{Comparison between the ratio-penalty and the three baseline selection methods for the ATIS dataset}
    \label{fig:resati}
\end{figure}

Figures \ref{fig:resres}, \ref{fig:resmov} and \ref{fig:resati} show the resulting curves of the three baselines and the ratio-penalty selection (RPS) for the MIT Restaurant, the MIT Movie and the ATIS dataset, respectively. The x-axis shows the number of samples used for training and the y-axis shows the best F1 score obtained during training on a separate test set. As the three baselines rely on random sampling, we repeat the procedure five times and plot the mean together with a $90\%$ confidence interval. As expected, the classic active learning (ALC) algorithm performs poorly because it tends to choose uninformative samples at the boundary of the cloud. The randomized active learning (ALR) gives a much better score, but surprisingly, it remains comparable to the random data selection strategy. Finally, the ratio-penalty selection (RPS) yields the best results, outperforming the baselines by a significant margin across all three domains. For example, in the MIT Restaurant domain, the average gap between RPS and the best performing baseline, ALR, is around $6$ points in F1 score. The best result is obtained for $k=10$ samples, where we observe approximately $55\%$ relative improvement of RPS over ALR. Both in the MIT Restaurant and MIT Movie domains, RPS needs, on average, $20$ labeled sentences \textit{less} to match the performance of ALR.
\vspace{0.2cm}

\begin{figure}
    \centering
    \includegraphics[width=\linewidth]{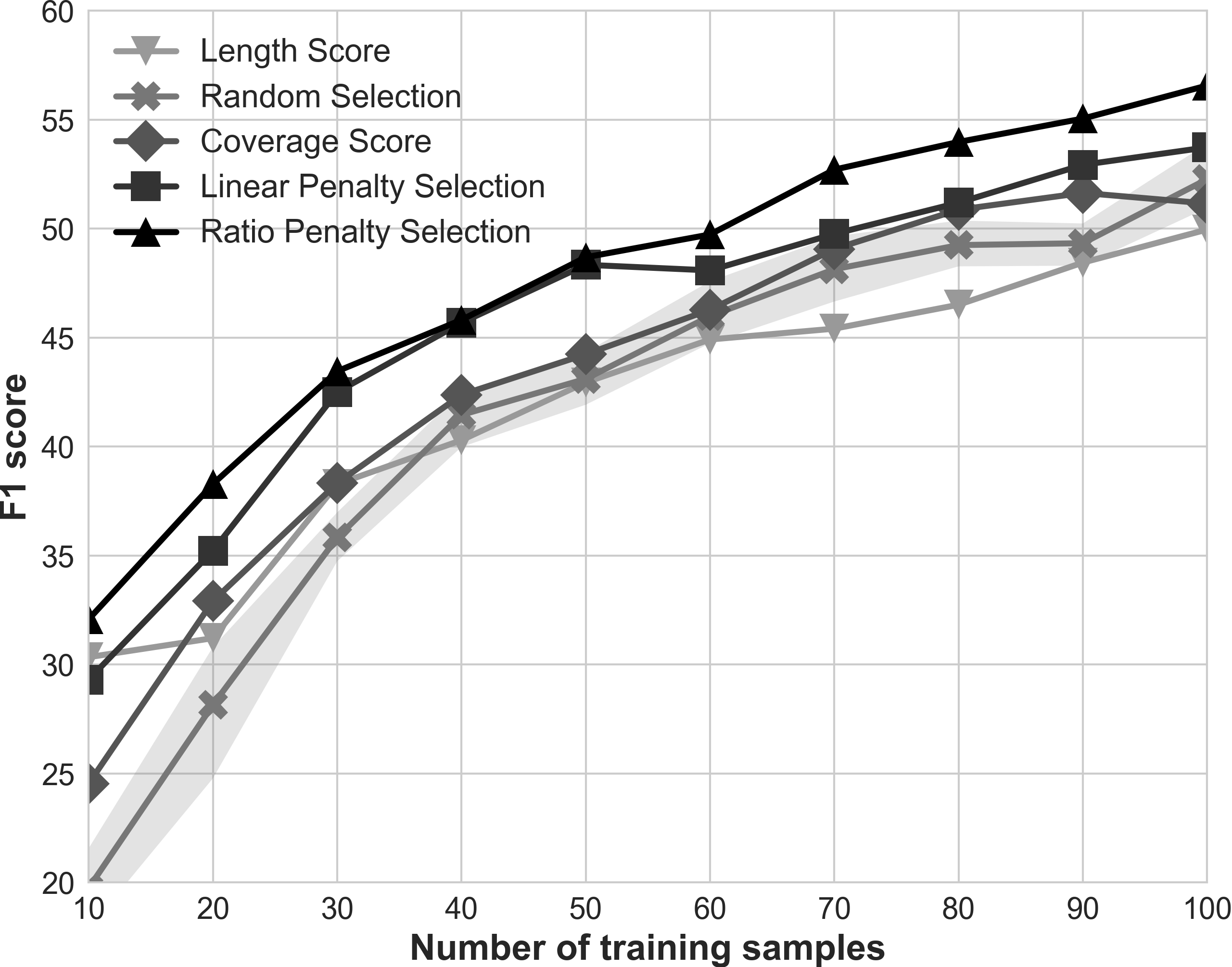}
    \caption{Comparison of various selection techniques applied on the MIT Restaurant dataset}
    \label{fig:resresall}
\end{figure}

 Figure \ref{fig:resresall} presents the resulting curves of different selection strategies for the MIT Restaurant dataset. The results were similar for the remaining two domains. The \textit{linear penalty} selection, introduced in Section \ref{sec:cs} and shown only for the best parameter $\alpha$, yields results that are better than the random choice and, in some regions, comparable to RPS. However, the disadvantage of this method is the requirement to tune an additional hyper-parameter that differs from one domain to another. We also present the performance when we train the model with the top $k$ \textit{longest} sentences (Length Score). It is fairly well in the very low data regimes, but it soon starts to degrade, becoming worse than all the other methods. 

\section{Conclusion}
In this paper, we have explored the utility of existing data selection approaches in the scenario of slot filling. We have introduced a novel submodularity-inspired selection technique, showing that a good choice of selection criteria can have a strong influence on the model's performance in the low-data regime. Moreover, we have shown that it is not necessary to limit this search to properly defined submodular functions. As they are efficiently optimized in practice by using the derived maximal gain, defining only the latter is sufficient to produce the rankings of the sentences. In addition, we have also defined a similarity metric between pairs of sentences using their continuous vector representations produced by a recent novel technique, \texttt{sent2vec}. Finally, we have shown that the space of raw samples already contains much information that can be exploited. This information is potentially more useful than the output space information used in the active learning paradigm.

\section*{Acknowledgements}
We would like to thank Patrick Thiran (EPFL), for sharing his valuable ideas and insights during the course of this research.

\appendix

\section{Adopted Model Specifications}
Our adopted model uses pre-trained GloVe embeddings on both word ($d_w=300$) and character level ($d_c=50$). The final word representation is a concatenation of its GloVe embedding and its character representation obtained by a separate mini-BiLSTM network. The hidden layer sizes of the main and the character BiLSTM are $h_w=200$ and $h_c=100$, respectively. Both the word and the character embeddings are fine-tuned during the training phase. Gradients are clipped to a maximum norm of $5$ and the learning rate of the Adam optimizer, which starts at $0.005$, is geometrically decayed at every epoch by a factor of $0.95$, until it reaches the minimum set value of $0.001$. Mini-batch size is equal to $20$.


\bibliographystyle{named}
\bibliography{main}

\begin{thebibliography}{}

\bibitem[\protect\citeauthoryear{Angeli \bgroup \em et al.\egroup
  }{2014}]{angeli2014stanford}
Gabor Angeli, Sonal Gupta, Melvin Jose, Christopher~D Manning, Christopher
  R{\'e}, Julie Tibshirani, Jean~Y Wu, Sen Wu, and Ce~Zhang.
\newblock Stanford’s 2014 slot filling systems.
\newblock {\em TAC KBP}, 695, 2014.

\bibitem[\protect\citeauthoryear{Bahdanau \bgroup \em et al.\egroup
  }{2014}]{bahdanau2014neural}
Dzmitry Bahdanau, Kyunghyun Cho, and Yoshua Bengio.
\newblock Neural machine translation by jointly learning to align and
  translate.
\newblock {\em arXiv preprint arXiv:1409.0473}, 2014.

\bibitem[\protect\citeauthoryear{Culotta and
  McCallum}{2005}]{culotta2005reducing}
Aron Culotta and Andrew McCallum.
\newblock Reducing labeling effort for structured prediction tasks.
\newblock In {\em AAAI}, volume~5, pages 746--751, 2005.

\bibitem[\protect\citeauthoryear{Fu \bgroup \em et al.\egroup
  }{2013}]{fu2013survey}
Yifan Fu, Xingquan Zhu, and Bin Li.
\newblock A survey on instance selection for active learning.
\newblock {\em Knowledge and information systems}, pages 1--35, 2013.

\bibitem[\protect\citeauthoryear{Hakkani-T{\"u}r \bgroup \em et al.\egroup
  }{2016}]{hakkani2016multi}
Dilek Hakkani-T{\"u}r, G{\"o}khan T{\"u}r, Asli Celikyilmaz, Yun-Nung Chen,
  Jianfeng Gao, Li~Deng, and Ye-Yi Wang.
\newblock Multi-domain joint semantic frame parsing using bi-directional
  rnn-lstm.
\newblock In {\em INTERSPEECH}, pages 715--719, 2016.

\bibitem[\protect\citeauthoryear{Huang \bgroup \em et al.\egroup
  }{2015}]{huang2015bidirectional}
Zhiheng Huang, Wei Xu, and Kai Yu.
\newblock Bidirectional lstm-crf models for sequence tagging.
\newblock {\em arXiv preprint arXiv:1508.01991}, 2015.

\bibitem[\protect\citeauthoryear{Jaech \bgroup \em et al.\egroup
  }{2016}]{jaech2016domain}
Aaron Jaech, Larry Heck, and Mari Ostendorf.
\newblock Domain adaptation of recurrent neural networks for natural language
  understanding.
\newblock {\em arXiv preprint arXiv:1604.00117}, 2016.

\bibitem[\protect\citeauthoryear{Kirchhoff and
  Bilmes}{2014}]{kirchhoff2014submodularity}
Katrin Kirchhoff and Jeff Bilmes.
\newblock Submodularity for data selection in statistical machine translation.
\newblock In {\em 2014 Conference on Empirical Methods in Natural Language
  Processing (EMNLP)}, pages 131--141, 2014.

\bibitem[\protect\citeauthoryear{Krause and
  Golovin}{2014}]{krause2014submodular}
Andreas Krause and Daniel Golovin.
\newblock Submodular function maximization., 2014.

\bibitem[\protect\citeauthoryear{Kurata \bgroup \em et al.\egroup
  }{2016}]{kurata2016leveraging}
Gakuto Kurata, Bing Xiang, Bowen Zhou, and Mo~Yu.
\newblock Leveraging sentence-level information with encoder lstm for semantic
  slot filling.
\newblock {\em arXiv preprint arXiv:1601.01530}, 2016.

\bibitem[\protect\citeauthoryear{Lample \bgroup \em et al.\egroup
  }{2016}]{lample2016neural}
Guillaume Lample, Miguel Ballesteros, Sandeep Subramanian, Kazuya Kawakami, and
  Chris Dyer.
\newblock Neural architectures for named entity recognition.
\newblock {\em arXiv preprint arXiv:1603.01360}, 2016.

\bibitem[\protect\citeauthoryear{Lin and Bilmes}{2011}]{lin2011class}
Hui Lin and Jeff Bilmes.
\newblock A class of submodular functions for document summarization.
\newblock In {\em Proceedings of the 49th Annual Meeting of the Association for
  Computational Linguistics: Human Language Technologies-Volume 1}, pages
  510--520. Association for Computational Linguistics, 2011.

\bibitem[\protect\citeauthoryear{Lin and Bilmes}{2012}]{lin2012learning}
Hui Lin and Jeff~A Bilmes.
\newblock Learning mixtures of submodular shells with application to document
  summarization.
\newblock {\em arXiv preprint arXiv:1210.4871}, 2012.

\bibitem[\protect\citeauthoryear{Liu and Lane}{2016}]{liu2016attention}
Bing Liu and Ian Lane.
\newblock Attention-based recurrent neural network models for joint intent
  detection and slot filling.
\newblock {\em arXiv preprint arXiv:1609.01454}, 2016.

\bibitem[\protect\citeauthoryear{Liu \bgroup \em et al.\egroup
  }{2013a}]{liu2013asgard}
Jingjing Liu, Panupong Pasupat, Scott Cyphers, and Jim Glass.
\newblock Asgard: A portable architecture for multilingual dialogue systems.
\newblock In {\em Acoustics, Speech and Signal Processing (ICASSP), 2013 IEEE
  International Conference on}, pages 8386--8390. IEEE, 2013.

\bibitem[\protect\citeauthoryear{Liu \bgroup \em et al.\egroup
  }{2013b}]{liu2013query}
Jingjing Liu, Panupong Pasupat, Yining Wang, Scott Cyphers, and Jim Glass.
\newblock Query understanding enhanced by hierarchical parsing structures.
\newblock In {\em Automatic Speech Recognition and Understanding (ASRU), 2013
  IEEE Workshop on}, pages 72--77. IEEE, 2013.

\bibitem[\protect\citeauthoryear{Ma and Hovy}{2016}]{ma2016end}
Xuezhe Ma and Eduard Hovy.
\newblock End-to-end sequence labeling via bi-directional lstm-cnns-crf.
\newblock {\em arXiv preprint arXiv:1603.01354}, 2016.

\bibitem[\protect\citeauthoryear{McCallum \bgroup \em et al.\egroup
  }{1998}]{mccallum1998employing}
Andrew McCallum, Kamal Nigam, et~al.
\newblock Employing em and pool-based active learning for text classification.
\newblock In {\em ICML}, volume~98, pages 350--358, 1998.

\bibitem[\protect\citeauthoryear{Mesnil \bgroup \em et al.\egroup
  }{2015}]{mesnil2015using}
Gr{\'e}goire Mesnil, Yann Dauphin, Kaisheng Yao, Yoshua Bengio, Li~Deng, Dilek
  Hakkani-Tur, Xiaodong He, Larry Heck, Gokhan Tur, Dong Yu, et~al.
\newblock Using recurrent neural networks for slot filling in spoken language
  understanding.
\newblock {\em IEEE/ACM Transactions on Audio, Speech and Language Processing
  (TASLP)}, 23(3):530--539, 2015.

\bibitem[\protect\citeauthoryear{Nemhauser \bgroup \em et al.\egroup
  }{1978}]{nemhauser1978analysis}
George~L Nemhauser, Laurence~A Wolsey, and Marshall~L Fisher.
\newblock An analysis of approximations for maximizing submodular set
  functions—i.
\newblock {\em Mathematical Programming}, 14(1):265--294, 1978.

\bibitem[\protect\citeauthoryear{Pagliardini \bgroup \em et al.\egroup
  }{2017}]{pagliardini2017unsupervised}
Matteo Pagliardini, Prakhar Gupta, and Martin Jaggi.
\newblock Unsupervised learning of sentence embeddings using compositional
  n-gram features.
\newblock {\em arXiv preprint arXiv:1703.02507}, 2017.

\bibitem[\protect\citeauthoryear{Wei \bgroup \em et al.\egroup
  }{2014}]{wei2014submodular}
Kai Wei, Yuzong Liu, Katrin Kirchhoff, Chris Bartels, and Jeff Bilmes.
\newblock Submodular subset selection for large-scale speech training data.
\newblock In {\em Acoustics, Speech and Signal Processing (ICASSP), 2014 IEEE
  International Conference on}, pages 3311--3315. IEEE, 2014.

\bibitem[\protect\citeauthoryear{Wei \bgroup \em et al.\egroup
  }{2015}]{wei2015submodularity}
Kai Wei, Rishabh Iyer, and Jeff Bilmes.
\newblock Submodularity in data subset selection and active learning.
\newblock In {\em International Conference on Machine Learning}, pages
  1954--1963, 2015.

\bibitem[\protect\citeauthoryear{Zhai \bgroup \em et al.\egroup
  }{2017}]{zhai2017neural}
Feifei Zhai, Saloni Potdar, Bing Xiang, and Bowen Zhou.
\newblock Neural models for sequence chunking.
\newblock In {\em AAAI}, pages 3365--3371, 2017.

\bibitem[\protect\citeauthoryear{Zhang and Wang}{2016}]{zhang2016joint}
Xiaodong Zhang and Houfeng Wang.
\newblock A joint model of intent determination and slot filling for spoken
  language understanding.
\newblock In {\em IJCAI}, pages 2993--2999, 2016.

\bibitem[\protect\citeauthoryear{Zhu and Yu}{2017}]{zhu2017encoder}
Su~Zhu and Kai Yu.
\newblock Encoder-decoder with focus-mechanism for sequence labelling based
  spoken language understanding.
\newblock In {\em Acoustics, Speech and Signal Processing (ICASSP), 2017 IEEE
  International Conference on}, pages 5675--5679. IEEE, 2017.

\end{thebibliography}

\end{document}